\documentclass[10pt,twocolumn,letterpaper]{article}

\usepackage{iccv}
\usepackage{times}
\usepackage{epsfig}
\usepackage{graphicx}
\usepackage{amsmath}
\usepackage{amssymb}
\usepackage{subfig}
\usepackage{comment}
\usepackage{multirow}
\usepackage{tabulary}

\usepackage[pagebackref=true,breaklinks=true,letterpaper=true,colorlinks,bookmarks=false]{hyperref}

\iccvfinalcopy 

\def\iccvPaperID{8242} 

\begin{document}

\title{Achieving Domain Robustness in Stereo Matching Networks by Removing Shortcut Learning}

\author{WeiQin Chuah, Ruwan Tennakoon,
        and~Alireza~Bab-Hadiashar\\
RMIT University, Australia\\
{\tt\small \{wei.qin.chuah,ruwan.tennakoon,abh\}@rmit.edu.au}
\and
David Suter\\
Edith Cowan University~(ECU), Australia\\
{\tt\small d.suter@ecu.edu.au}
}
\maketitle
\ificcvfinal\thispagestyle{empty}\fi

\begin{abstract}
   Learning-based stereo matching and depth estimation networks currently excel on public benchmarks with impressive results. However, state-of-the-art networks often fail to generalize from a synthetic imagery to more challenging real data domains. This paper is an attempt to uncover hidden secrets of achieving domain robustness and in particular, discovering the important ingredients of generalization success of stereo matching networks by analyzing the effect of synthetic image learning on real data performance. We provide evidence that demonstrates that learning of features in synthetic domain by a stereo matching network is heavily influenced by two ``shortcuts" presented in the synthetic data: {\normalfont (1)}~identical local statistics~(RGB colour features) between matching pixels in the synthetic stereo images and {\normalfont (2)}~lack of realism in synthetic textures on 3D objects simulated in game engines.
   We will show that by removing such shortcuts, we can achieve domain robustness in the state-of-the-art stereo matching frameworks and produce remarkable performance on multiple realistic datasets, despite the fact that the networks were trained on synthetic data, only. Our experimental results point to the fact that eliminating shortcuts from the synthetic data is key to achieve domain-invariant generalization between synthetic and real data domains.
   
\end{abstract}

\section{Introduction} \label{Sec:Intro}
Stereo matching is a fundamental problem in computer vision and is widely used in various applications such as augmented reality~(AR), robotics and autonomous driving. Stereo matching aims to estimate depth by computing the horizontal displacement of pixel correspondences between a pair of stereo images. In recent years, many end-to-end Convolutional Neural Networks~(CNNs) have been developed to perform stereo matching and achieved outstanding results on several publicly available datasets or benchmarks~\cite{chang2018pyramid, guo2019group, kendall2017end, xu2020aanet, zhang2019ga}. In practice, the state-of-the-art stereo matching networks are trained in a supervised fashion where annotated datasets are required to fine-tune the models from synthetic to real data domains. However, the ground-truth disparity labels are cumbersome to generate in real-world scenarios.

\begin{figure}
    \centering
    \subfloat[Left Image]{\includegraphics[width=0.15\textwidth]{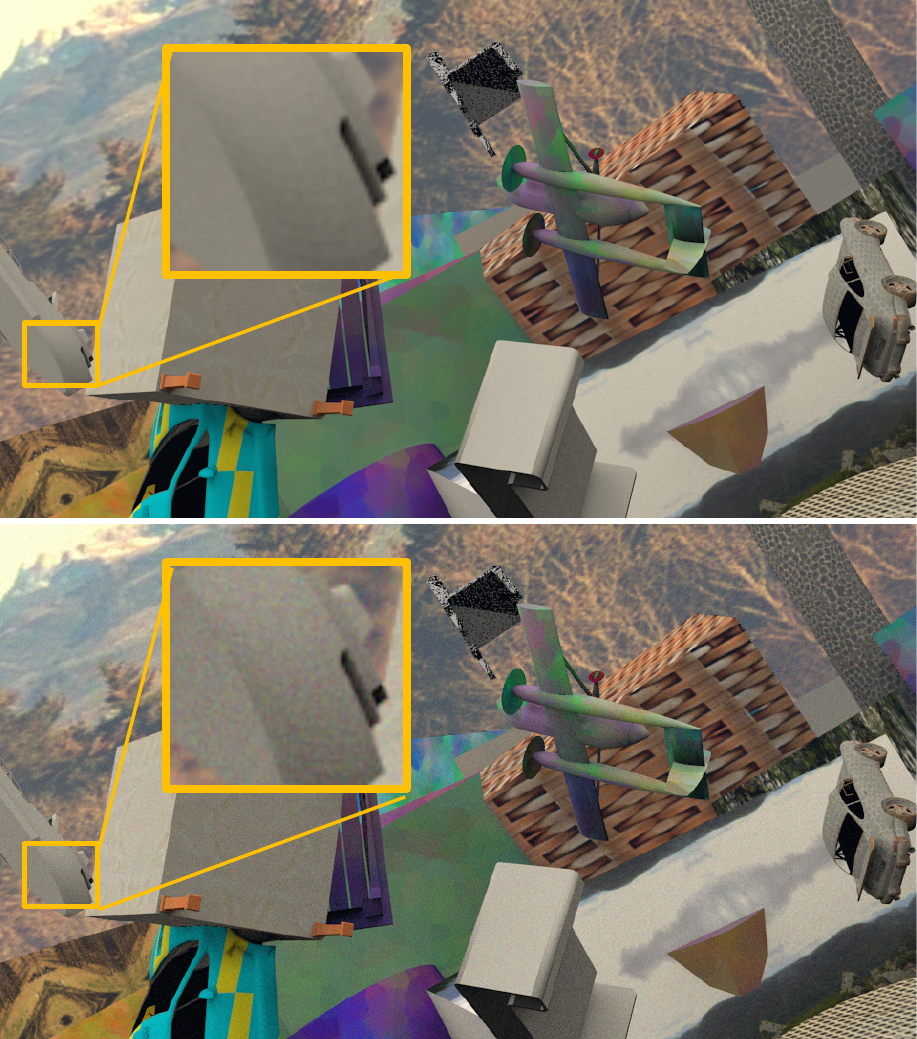}}  
    \hspace{.5mm}
    \subfloat[Baseline]{\includegraphics[width=0.15\textwidth]{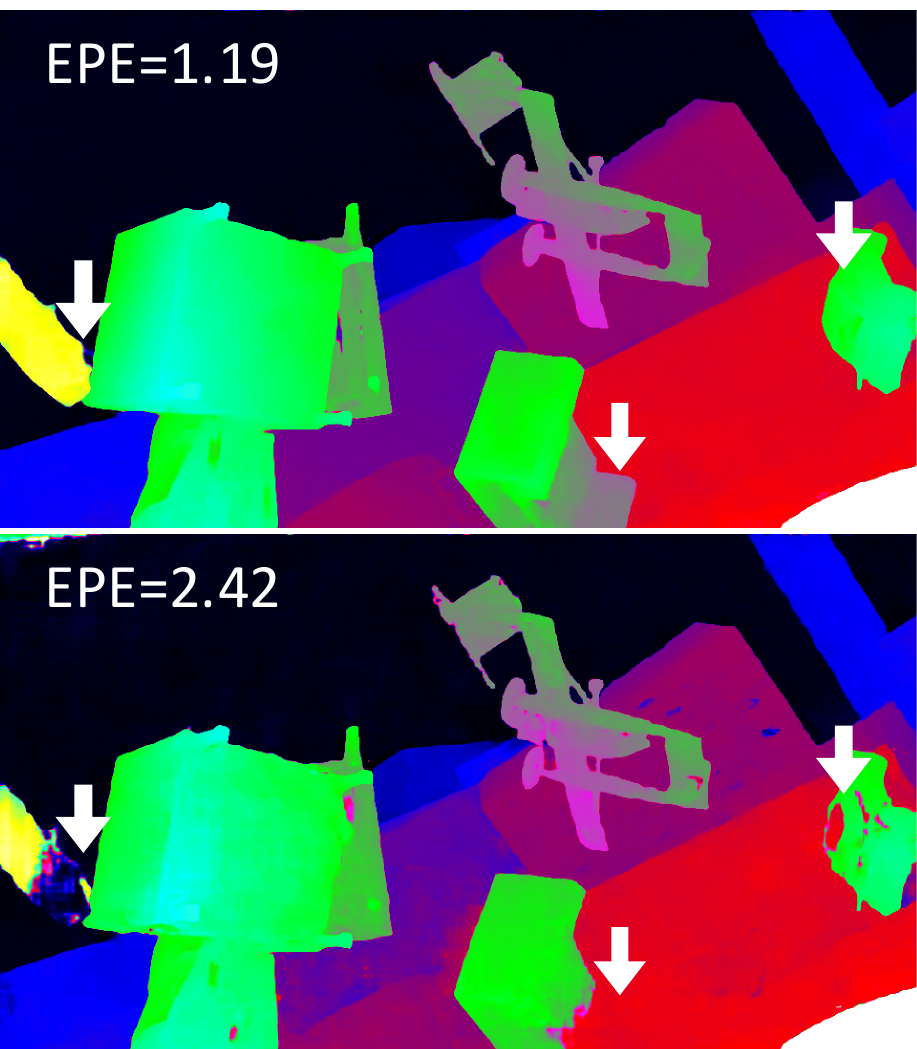}}
    \hspace{.5mm}
    \subfloat[Ours]{\includegraphics[width=0.15\textwidth]{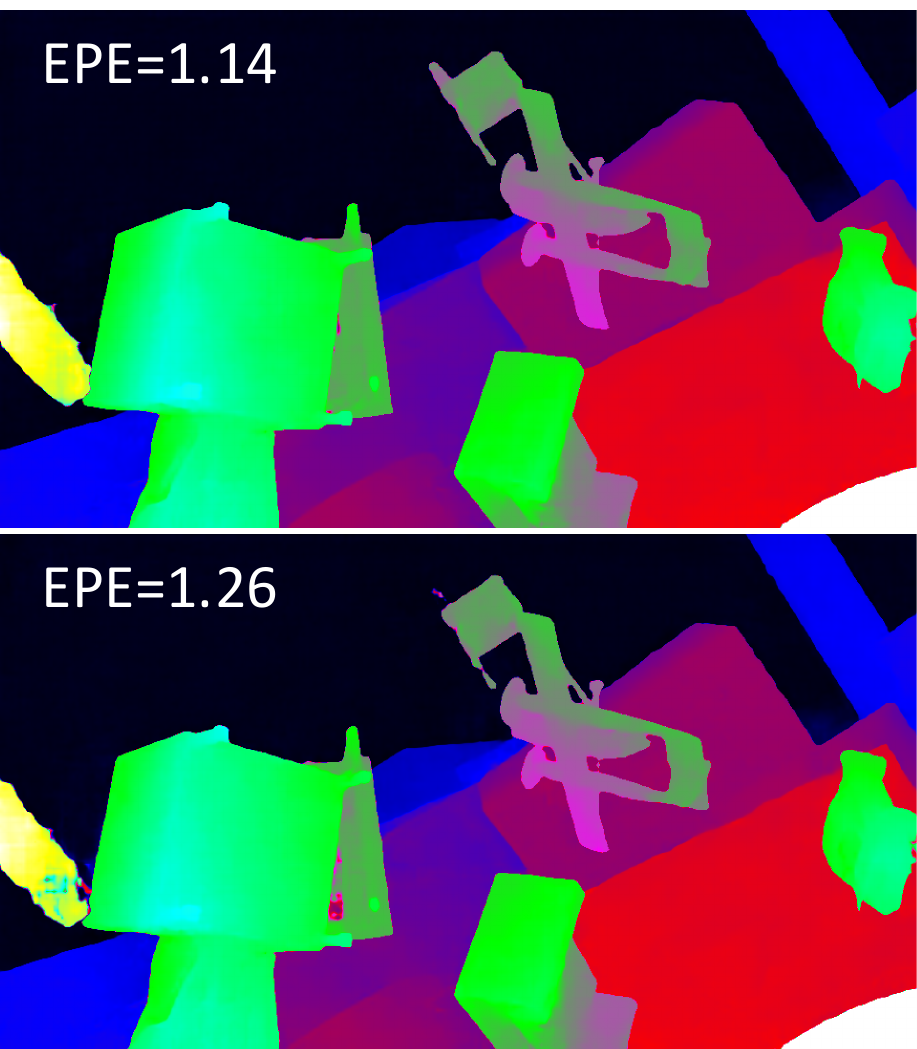}}
    \caption{(Best view in color and zoom in)~Performance comparison between stereo matching networks with~(baseline) and without the shortcuts removed. The performance of the baseline network deteriorated when adversarial noises that is hardly visible to human eyes are added to the stereo image~(bottom).} 
    \label{fig:noise}
\end{figure}

A major drawback of the existing learning-based stereo matching networks is their inability to generalize to unseen domains. It is commonly understood that this is due to domain differences between the training and testing data~\cite{wang2018deep}. The differences may include discrepancies in image appearance, style and contents between datasets. To overcome this, unsupervised domain adaptation~(UDA) methods were proposed to bridge the domain gaps between synthetic and real data, and to effectively transfer learned knowledge without relying on ground truth in the target domain~\cite{liu2020stereogan, tonioni2017unsupervised, tonioni2019learning, tonioni2019real}. Yet, UDA requires a large set of stereo images from the target domain, which is challenging to acquire in many real-world scenarios.

Conversely, domain generalization~(DG) allows the network to learn domain-invariant features without requiring specific information of the target domain~\cite{hu2020domain}. For instance, Zhang~\etal~\cite{zhang2020domain} proposed to regularize the distribution of the extracted features using domain normalization to attain domain-invariant representation. 


\begin{figure*}[th]
    \centering
    \includegraphics[width=0.21\textwidth]{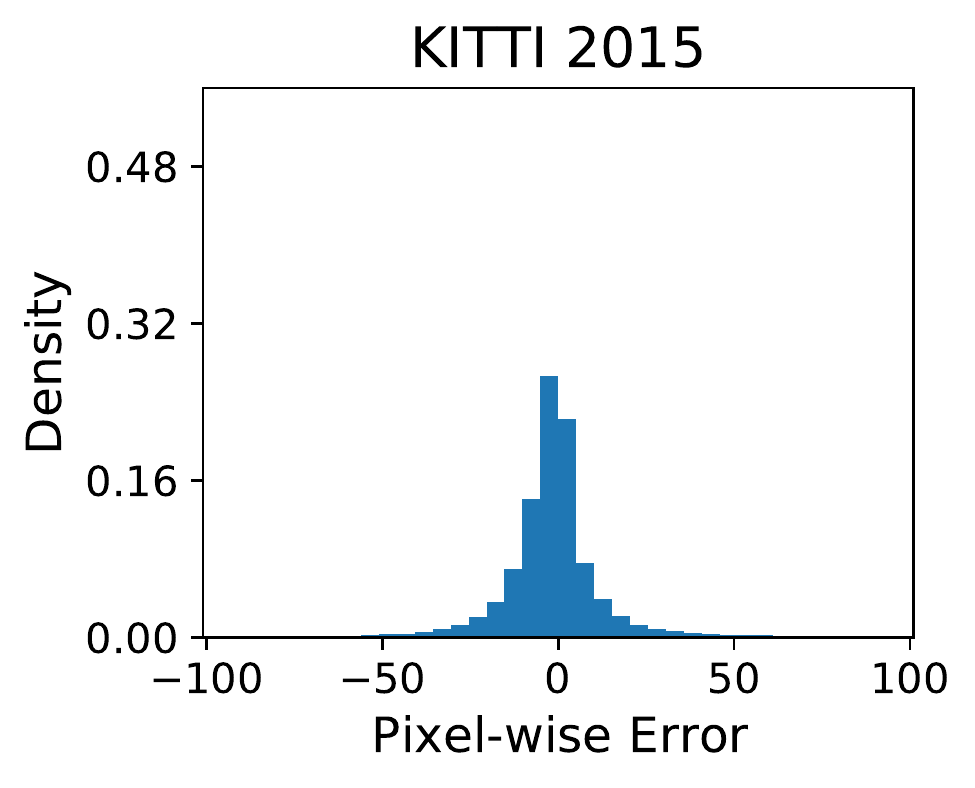} 
    \hspace{1.5mm}
    \includegraphics[width=0.21\textwidth]{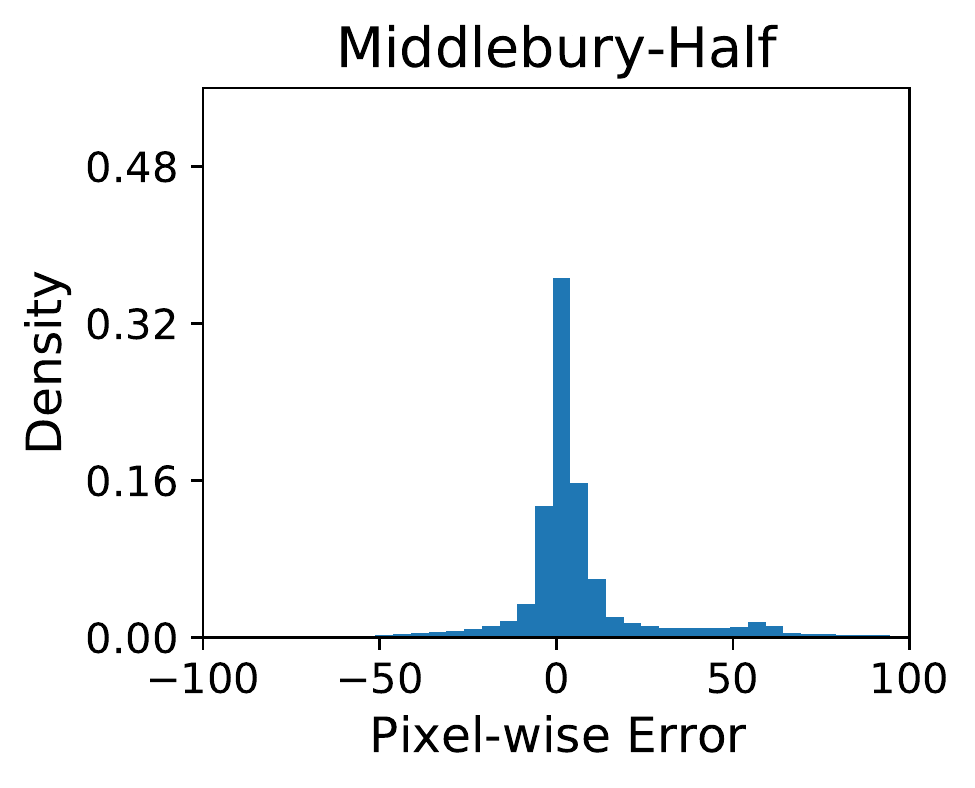}
    \hspace{1.5mm}
    \includegraphics[width=0.21\textwidth]{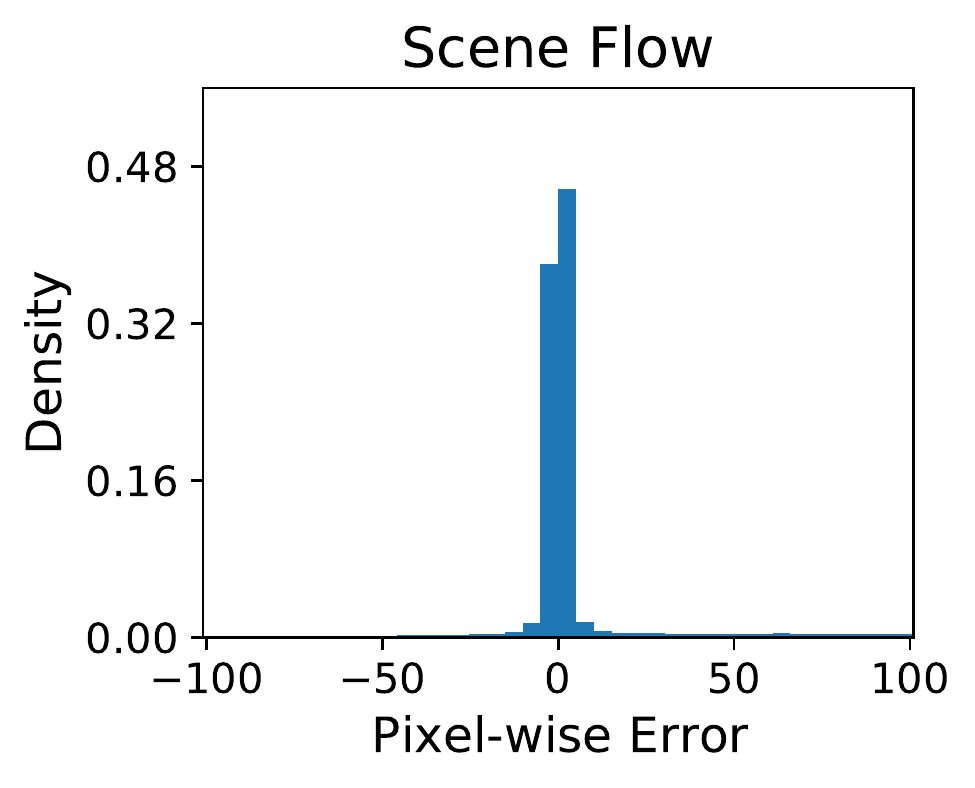}
    \hspace{1.5mm}
    \includegraphics[width=0.21\textwidth]{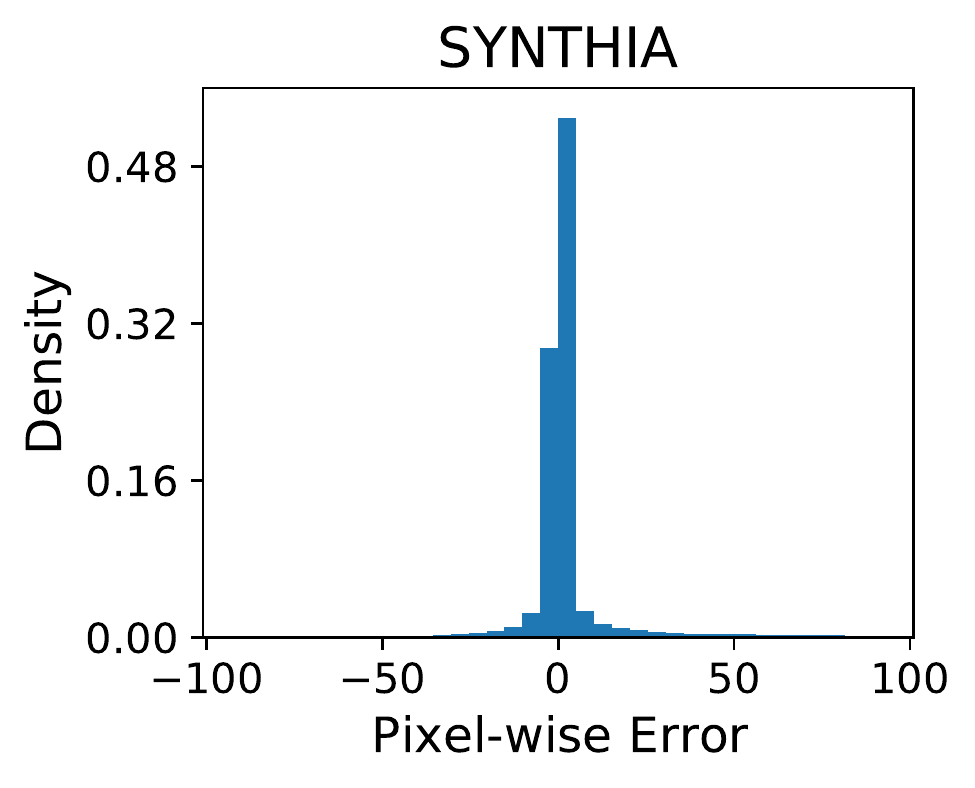}
    \caption{Color discrepancy in RGB channels between pixel correspondences in different datasets. Significant portion of pixels have greater color discrepancy in real datasets~(KITTI~2015, Middlebury) as compared to synthetic datasets~(SceneFlow, SYNTHIA).}
    \label{fig:color_error}
\end{figure*}

It is important to note that both methods (i.e. UDA and DG) are designed to mitigate the effect of discrepancies between synthetic and real data domains. Here, we argue that the main issue that prevents stereo matching networks from generalizing across domains is the learning of trivial features in the synthetic domains. Geirhos~\etal~\cite{geirhos2020}~coined a useful term for this phenomenon as ``shortcut learning", where ``shortcut" represents a solution that attains excellent performance on data similar to the training environment but fails to generalize to more challenging testing conditions, such as real-world scenarios. We have identified that the occurrence of shortcut learning in stereo matching network is mainly induced by the natural attributes of synthetic stereo images, including: (1)~identical local statistics~(RGB colour features) between matching pixels in the synthetic stereo images and (2)~lack of realism in synthetic textures included on 3D object models. 


We found that the identical color features between pixel correspondences in synthetic stereo images provide the stereo matching networks with easy hints to locate the matching pixels. As shown in Fig.~\ref{fig:color_error}, a substantial portion of pixels in the synthetic stereo images have a lower color discrepancy between correspondences as compared to realistic indoor or outdoor stereo images. Consequently, the network trained on synthetic data is diverted from learning the intended robust and domain-invariant features, as trivial features are sufficient to accomplish (the delusion of) superior performance in the synthetic domain. As such, a stereo matching network trained on synthetic data is highly susceptible to this color hint, and fails drastically when tested on synthetic data with insignificant perturbation included in the color distribution~(shown in Fig.~\ref{fig:noise}). In its supplementary material, this paper provides empirical evidence, illustrating that the solution provided by the stereo matching network trained on synthetic data is highly susceptible to this color hint. 

Furthermore, we discover that stereo matching networks are biased to exploit trivial and local features to estimate disparity for synthetic objects~(e.g. cars). This is mainly due to the lack of realistic textures on synthetic objects, reducing the difficulty in estimating matching pixels between stereo views. However, complex high-level features~(shape, semantics) that are robust to domain changes are highly desirable to improve generalization~\cite{geirhos2018imagenet}. In practice, this issue is often mitigated by performing image-to-image translation using a generative adversarial model (GANs) to close the gap between synthetic images and the realistic target domain. However, it is challenging to adopt such a method in stereo matching as the generative model does not guarantee epipolar consistency, nor feature consistency between stereo views~\cite{liu2020stereogan}.

This paper aims to demonstrate that domain-robustness can be achieved in stereo matching networks by removing the mentioned shortcuts from the synthetic training data. To this end, we propose to include two existing data augmentation techniques, namely asymmetric chromatic augmentation and asymmetric random patching, to remove these shortcuts from synthetic stereo images. It is worth noting that a better approach may exist to eliminate the mentioned shortcuts than the included data augmentation methods. However, the focus of this work is to illustrate that the removal of shortcuts can lead to remarkable improvement in domain-invariant generalization in stereo matching networks, despite using a simple method like the included augmentation techniques. 
Our experimental results suggest that eliminating the shortcut features in synthetic stereo images is key to attaining domain-invariant stereo matching networks. 
Our implementation is available at: {URL}.

\noindent In summary, our main contributions include: 
\vspace{-2mm}
\begin{itemize}
    \item demonstrating that the shortcuts found in synthetic stereo images lead to the learning of domain specific and trivial features in stereo matching networks;\vspace{-2mm}
    \item showing that eliminating the identified shortcuts from synthetic data using data augmentation can attain domain-robustness in stereo matching networks;\vspace{-2mm}
    \item achieving impressive performance in stereo matching and disparity estimation in multiple realistic domains using stereo matching network trained on synthetic data only; and\vspace{-2mm} 
    \item provision of a simple blueprint as a significant step to design a domain-invariant stereo matching networks without requiring sophisticated network alternation or additional learnable parameters. 
\end{itemize}

\section{Related Work}

\noindent \textbf{Shortcut Learning} \\
Shortcuts have been described as decision rules that involve using trivial features to achieve superior performance on independent and identically distributed~(i.i.d) test data but fail on out-of-distribution~(o.o.d) test data~\cite{geirhos2020}. The occurrence of shortcuts in deep neural networks~(DNNs) are mainly due to the shortcut opportunities presented in data~(dataset bias) and the selective feature combination. In image classification, DNNs usually learn the unintended solutions by leveraging the systematic relationship between object and background or context to correctly label the image. Hence, DNNs fail when the commonly seen context is changed or removed. For example, a cow that is located on a beach is not classified correctly as cows are usually found on grass field~\cite{Beery_2018_ECCV}. Moreover, DNNs are also biased to the extraction of trivial features which are specialized to related i.i.d test data. For example, DNNs trained on ImageNet dataset for image classification are biased toward utilizing texture cue to recognize objects despite that shape cue promotes robustness towards different image distortion~\cite{geirhos2018imagenet}. 

Similarly, stereo matching networks have a tendency of exploiting shortcuts when they achieve impressive results on synthetic data. However, those fail to generalize to realistic domains. A compounding factor is that shortcuts can be unintuitive and difficult to recognize~\cite{minderer2020automatic}. In this paper, we attempt to identify and formalize the underlying shortcuts that are preventing the stereo matching networks from attaining domain robustness. Our experimental results demonstrate that eliminating the identified shortcuts can significantly improve domain robustness in most stereo matching networks. \\

\noindent \textbf{Learning-based Stereo Matching Networks} \\
In recent years, end-to-end learning deep stereo matching networks have accomplished significant success and excel in most datasets and benchmarks~\cite{chang2018pyramid, kendall2017end, xu2020aanet, zhang2019ga}. These networks can be categorized into two groups: (1) correlation-based stereo matching networks and (2) concatenation-based stereo matching networks. 

The correlation-based stereo matching networks were first proposed in DispNetC~\cite{mayer2016large}. They construct the similarity cost volume by correlating deep features extracted from the stereo views. The networks also learn to predict dense disparity map by minimizing a disparity-based regression loss function~(e.g. L1 loss). Other state-of-the-art correlation-based stereo matching networks include iResNet~\cite{liang2018learning}, CRL~\cite{pang2017cascade}, SegStereo~\cite{yang2018segstereo}, EdgeStereo~\cite{song2018edgestereo} and AANet~\cite{xu2020aanet}. For example, iResNet~\cite{liang2018learning} and CRL~\cite{pang2017cascade} are designed to rectify the initially predicted disparity map by adding residual signals generated by a subsequent network. On the other hand, SegStereo~\cite{yang2018segstereo} and EdgeStereo~\cite{song2018edgestereo} include multitask learning networks that combine stereo matching with an auxiliary task~(e.g. semantic segmentation, edge detection). Moreover, AANet~\cite{xu2020aanet} uses an adaptive multi-scale cost aggregation method and a content-aware intra-scale cost aggregation method, designed using deformable convolution.

In contrast, concatenation-based stereo matching networks learn to estimate feature similarity directly from the stacked left and right features. Multiple 3D-CNNs stacked hourglass modules are included in the network for cost aggregation and regularization. Dissimilar to the correlation-based networks that directly estimate dense disparity maps~(except AANet), the concatenation-based networks generate dense disparity maps using the estimated cost volumes via soft-argmax~\cite{kendall2017end}. Examples of the state-of-the-art concatenation-based stereo matching networks include PSMNet~\cite{chang2018pyramid}, GANet~\cite{zhang2019ga}, GCNet~\cite{kendall2017end}, StereoNet~\cite{khamis2018stereonet}, StereoDrNet~\cite{chabra2019stereodrnet} and EMCUA~\cite{nie2019multi}.

While these networks have superior performance in stereo matching, labelled samples in target environments are mandatory for fine-tuning. Without fine-tuning, these networks cannot generalize to the new test data and the performance deteriorates drastically. Alternatively, self-supervised or unsupervised stereo matching networks eliminate the need for ground truth labels by employing unsupervised losses such as reconstruction loss, smoothness loss, structural similarity~(SSIM) loss and left-right consistency loss~\cite{aleotti2020reversing, li2018occlusion, zhong2017self, zhou2017unsupervised}. Yet, these networks require large number of training samples from the target domain and can hardly generalize to novel domains. In this work, we illustrate that stereo matching networks can be optimized on synthetic data only and attain desirable performance on challenging realistic data without fine-tuning, by eliminating the identified shortcuts. \\

\noindent \textbf{Unsupervised Domain Adaptation in Stereo Matching} \\
Unsupervised domain adaptation~(UDA) involves transferring learned knowledge from source to target domain without using ground truth labels. In the context of stereo matching, Tonioni~\etal~\cite{tonioni2017unsupervised} employed traditional stereo matching algorithms and confidence measures to generate reliable proxy labels, which are used to fine-tune the pre-trained networks. Conversely, in~\cite{tonioni2019learning} the adaptation procedure was formulated as part of the learning process, enabling the stereo matching networks to learn to adapt, using meta-learning. Also, Pang~\etal~\cite{Pang_2018_CVPR} proposed to use graph Laplacian regularization to iteratively optimize estimated disparities at multiple resolutions to adapt to new domains. Several online adaptation methods were also proposed for stereo matching networks~\cite{tonioni2019real, Zhong_2018_ECCV}. 

Although these methods can effectively generalize pre-trained stereo networks to novel domains, a large set of training images from the new domains is still required. As mentioned in section~\ref{Sec:Intro}, the required data collection process involves exhaustive efforts. \\

\noindent \textbf{Domain Generalization in DNNs} \\
Domain generalization allows DNNs to perform consistently well across different target domains while trained using data from the source domain only. This is often achieved by enabling the networks to learn domain-invariant features~\cite{li2018domain, li2018deep, muandet2013domain, qiao2020learning}. For instance, Li~\etal~\cite{li2018domain} proposed to align latent feature distribution in source and target domains (with an arbitrary prior distribution) by minimizing a distance-based metric~(e.g. Maximum Mean Discrepancy). Alternatively, Qiao~\etal~\cite{qiao2020learning} proposed to expand the training sets by generating fake and challenging examples via meta-learning based adversarial augmentation. 

In stereo matching, Zhang~\etal~\cite{zhang2020domain} proposed a novel Domain Normalization method to replace Batch or Instance Normalization, which allows the network to extract domain-invariant features. Additionally, they also proposed DSMNet, a graph-based filtering stereo matching network that utilizes the domain invariant features to preserve structural and geometric representation, which further promotes generalization across domains. 

In contrast, we have identified shortcut learning~\cite{geirhos2020} as a major factor that hinders stereo matching networks from generalizing across domains. To this end, we propose to remove the shortcut opportunities in the synthetic dataset, using data augmentation. Our method can be easily implemented in the training pipeline, without including any additional trainable parameters or changes to the network architecture. In addition, our results demonstrate that a large step towards domain-invariance in stereo matching networks can be taken by simply including the proposed data augmentation methods. 

\section{Methodology}
\begin{figure}[t]
    \centering
    \vskip -0.35cm
    \subfloat[Input Image]{\includegraphics[width=0.145\textwidth]{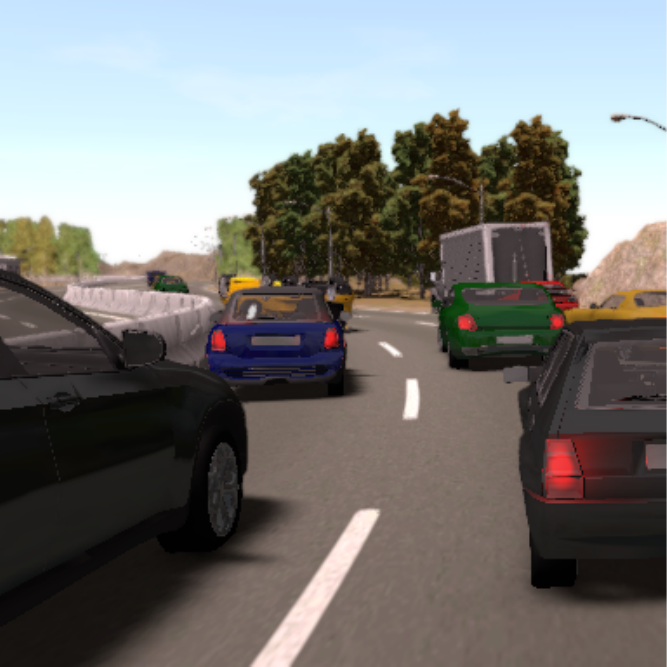}}
    \hspace{1mm}
    \subfloat[ACA]{\includegraphics[width=0.145\textwidth]{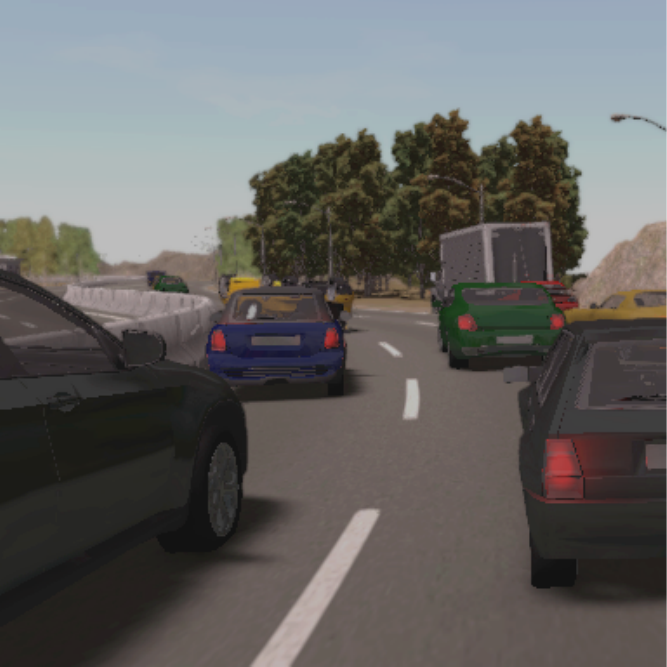}}
    \hspace{1mm}
    \subfloat[ACA + ARP]{\includegraphics[width=0.145\textwidth]{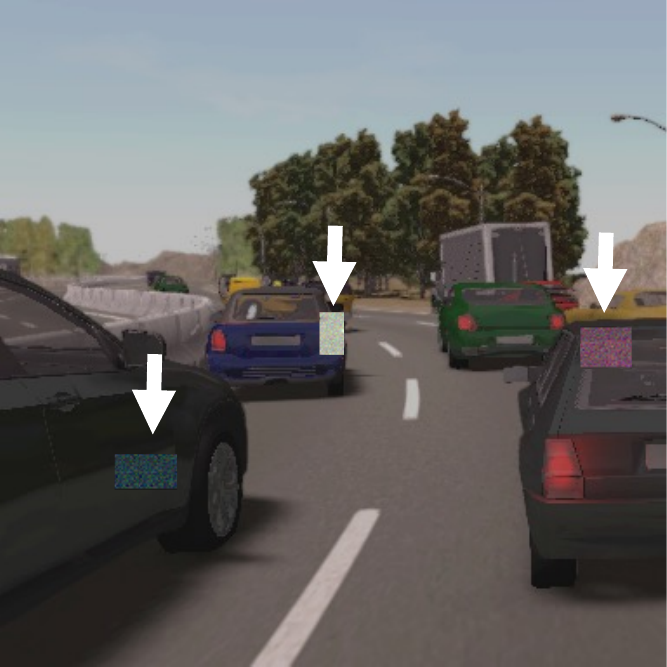}}
    \caption{(Best view in color and zoom in)~Example of a synthetic image sampled from SYNTHIA dataset~\cite{Ros_2016_CVPR} with the proposed data augmentation methods: asymmetric chromatic augmentation~(ACA) and asymmetric random patching~(ARP). \vspace{-2mm}}
    \label{fig:augment_example}
\end{figure}

In this section, we will discuss the intuition behind the selection of data augmentation methods employed in stereo methods to mitigate the effect of identified shortcuts. These augmentation methods will be combined with the commonly employed data pre-processing procedures in stereo matching networks, such as random cropping and image normalization. In addition, the outcomes of the data augmentation methods are illustrated in Fig.~\ref{fig:augment_example}. Details regarding experimental setup and results are provided in Section~\ref{Sec:Experiments}. \\

\noindent \textbf{Asymmetric Chromatic Augmentation} \\
Stereo matching networks trained on synthetic data fail to generalize to real domains, mainly due to the identical color features between matching pixels in synthetic stereo images as discussed in Sec.~\ref{Sec:Intro}. This results in the learning of simple solutions, such as matching trivial features~(colors, textures, etc.), which are sufficient to estimate disparity accurately in the synthetic domain. However, in practice, robust and high-level features are required~(to generalize to a more challenging domain~\cite{geirhos2020, geirhos2018imagenet}). To resolve this issue, we propose to eliminate the shortcut opportunity arising from the color hint, by removing the similarity in colors between pixel correspondences, using asymmetric chromatic augmentation. Specifically, the illumination and colors of left and right stereo images are changed differently by adjusting the brightness, contrast and saturation parameters within a pre-defined range. 

Although asymmetric chromatic augmentation was included in HSMNet~\cite{yang2019hierarchical} to mitigate the effect of varying lighting and exposure conditions under different stereo viewpoints, the benefit of this augmentation for robustness in terms of generalizing stereo matching networks was not previously discovered. In fact, they suggested that for mixed training (using synthetic and real data) of the network, better performance was attained without using this augmentation. We have however discovered that for achieving domain robustness in stereo matching networks and relying on the synthetic training only (a desirable option), the proper use of the asymmetric chromatic augmentation can effectively eliminate the color hint shortcut and significantly improve the performance~(refer to Table.~\ref{table:ablation}). In addition, we also discovered that asymmetric chromatic augmentation promotes illumination invariant as adjusting the brightness of the input image will generate a diverse set of additional training data with varying lighting conditions~(day time and night time). Following the HSMNet method, we randomly select the brightness parameters between the range of $[0.4, 2.0]$ and the saturation and contrast parameters between the range of $[0.5, 1.5]$. \\

\begin{figure}
    \centering
    \includegraphics[width=0.2\textwidth]{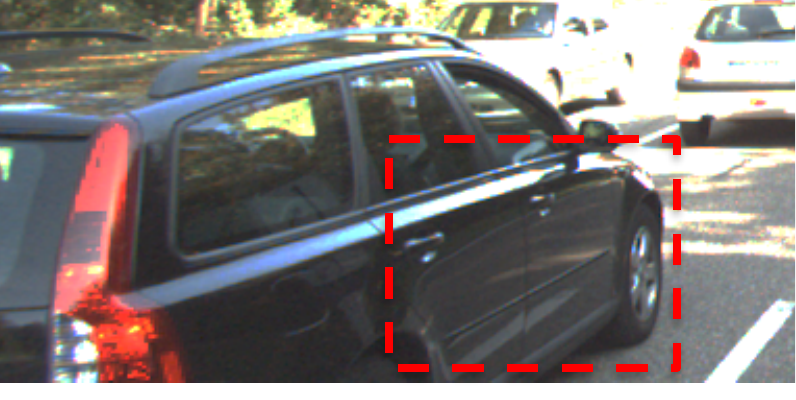}
    \hspace{1mm}
    \includegraphics[width=0.2\textwidth]{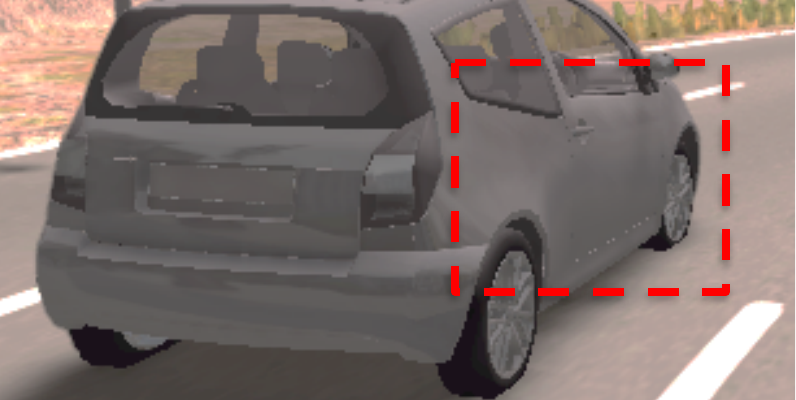}
    \caption{Qualitative comparison of a car object captured in the real world~(left) and a synthetic car model generated in simulation~(right). The synthetic car can be considered as an over-simplified model of the real car.}
    \label{fig:cars}
\end{figure}

\noindent \textbf{Asymmetric Random Patching} \\
In the publicly available synthetic stereo datasets~(e.g. SYNTHIA~\cite{Ros_2016_CVPR} and SceneFlow~\cite{mayer2016large}), the included 3D models are often overly-simplified versions of the real-world objects. For example, reflections on car doors presented in the realistic image are not modelled in the synthetic data as illustrated in Fig.~\ref{fig:cars}. The lack of realistic image features on these objects significantly simplifies the learning aspects of the synthetic domain. Consequently, stereo matching networks trained on synthetic stereo images will learn to exploit overly-simplistic local features that are specialized for synthetic domain. In contrast, features such as global contextual cues and robust structural representation of the observed objects are desirable for domain generalization~\cite{zhang2020domain, geirhos2018imagenet}. 

\begin{table}[t]
\centering
\resizebox{0.47\textwidth}{!}{%
\begin{tabular}{c|cc|cc|cc|c}
\hline
\multirow{2}{*}{Methods} & \multicolumn{2}{l|}{Augmentation} & \multicolumn{2}{c|}{KITTI} & \multicolumn{2}{c|}{Middlebury} & \multirow{2}{*}{ETH3D} \\
 & ACA & ARP & 2012 & 2015 & Half & Quarter &  \\ \hline
\multirow{3}{*}{PSMNet} &  &  & 29.3 & 28.0 & 33.9 & 20.1 & 12.5 \\
 &  & \checkmark & 7.5 & 8.1 & 20.3 & 16.4 & 10.3 \\
 & \checkmark &  & 4.0 & 4.5 & 12.3 & 8.6 & 9.8 \\
 & \checkmark & \checkmark & \textbf{3.6} & \textbf{4.0} & \textbf{10.4} & \textbf{7.9} & \textbf{8.4} \\ \hline
\multirow{3}{*}{GwcNet} &  &  & 11.5 & 11.3 & 25.5 & 12.7 & 9.9 \\
 &  & \checkmark & 6.3 & 6.8 & 16.4 & 11.2 & 8.7 \\
 & \checkmark &  & 3.9 & 4.3 & 12.2 & 7.7 & 7.5 \\
 & \checkmark & \checkmark & \textbf{3.7} & \textbf{3.8} & \textbf{9.1} & \textbf{6.1} & \textbf{5.5} \\ \hline
\end{tabular}%
}
\caption{Ablation study of the proposed data augmentation methods: Asymmetric Chromatic Augmentation~(ACA), Asymmetric Random Patching~(ARP). All models are trained using SceneFlow and SYNTHIA synthetic datasets and tested using three different real datasets.}
\label{table:ablation}
\end{table}

To this end, we propose to use the asymmetric random patching to mitigate the effect of shortcut learning caused by the lack of realistic image features on synthetic objects. The asymmetric random patching method is inspired by the limited context inpainting~(LCI) proposed in~\cite{jenni2020steering}. The LCI involves employing a generative model to inpaint a randomly positioned local patch using the pixel information from the patch boundary. As a result, the inpainted patch retains local statistics~(identical to the boundary pixels) only, and does not correlate to the global context of the image. Learning to discriminate LCI from a list of image transformations~(e.g. warping and rotation) as a pretext task allows the network to exploit global information from the images and improves its generalization ability for the subsequent task~(i.e. image classification).

Similar to the LCI, asymmetric random patching perturbs several local patches positioned randomly in the left or right image~(at a chance of $p=0.5$). These perturbations include changes in color, and addition of grainy noise sampled from a Gaussian distribution, $\mathcal{N}(\mu=0, \sigma=0.1)$. This increases the chance of the network to learn robust and descriptive contextual cues for objects and image context, promoting domain generalization. Furthermore, by including asymmetric random patching, the network learns to exploit monocular cues from either of the stereo views that is robust to occlusion. Consequently, the network is capable of estimating accurate disparity measurements even when the objects are partly occluded in either of the stereo viewpoints~(supporting evidences are provided as supplementary material). In our implementation, the number of local patches included in an image is sampled uniformly between $[2, 4]$ and the height and width of each patch is sampled uniformly between $[50, 100]$~pixels. 


\section{Experiments} \label{Sec:Experiments}
In this section, we evaluate the proposed method using seven datasets collected from outdoor and indoor realistic scenes. Importantly, all methods are trained only using synthetic data ~(SceneFlow~\cite{mayer2016large}, SYNTHIA~\cite{Ros_2016_CVPR}) and directly tested using realistic data~(KITTI2012~\cite{Geiger2012CVPR}, KITTI2015~\cite{Menze2018JPRS}, Middlebury~\cite{scharstein2014high}, DrivingStereo~\cite{yang2019drivingstereo} and ETH3D~\cite{schoeps2017cvpr}), without adaptation or fine-tuning. \\

\begin{table}[t]
\resizebox{0.47\textwidth}{!}{%
\begin{tabular}{c|c|c|c|c}
\hline
\multirow{2}{*}{Network} & \multirow{2}{*}{Normalization} & \multirow{2}{*}{Augmentation} & KITTI & Middlebury \\
 &  &  & 2015 & Half \\ \hline 
\multirow{4}{*}{PSMNet} & BN &  & 28.0 & 33.9 \\
 & DN &  & 6.6 & 18.1 \\
 & BN & ACA+ARP & \textbf{4.0} & \textbf{10.4}\\ \hline
\multirow{4}{*}{GwcNet} & BN &  & 11.3 & 25.5 \\
 & DN &  & 5.4 & 17.2 \\
 & BN & ACA+ARP & \textbf{3.8} & \textbf{9.1} \\ \hline
\end{tabular}%
}
\caption{Performance comparison with Domain Normalization~(DN)~\cite{zhang2020domain} in generalizing stereo matching networks. All models are trained using SceneFlow and SYNTHIA synthetic datasets.}
\label{table:DN}
\end{table}

\noindent \textbf{Datasets and Metrics: }
\textbf{KITTI2012} and \textbf{KITTI2015} provide 193 and 200 stereo images of outdoor driving scenes with sparse disparity ground truth in the training sets. \textbf{Middlebury} contains 15 images of high resolution indoor scenes with sparse ground truth. \textbf{ETH3D} provides 27 low resolution, greyscale stereo images with sparse ground truth. \textbf{DrivingStereo} is a large-scale real dataset, covering a diverse set of driving scenarios and different weather conditions; containing over 174,437 stereo pairs for training. We also test the robustness of our proposed method using data collected in different weather conditions~(provided by the DrivingStereo dataset).

\textbf{SceneFlow} is a large collection of synthetic stereo images with dense disparity ground truth. It contains three subsets with different settings: FlyingThings3D, Driving and Monkaa and provides 35,454 training and 4,370 testing images. \textbf{SYNTHIA} composes of 50 different video sequences rendered in different seasons and weather conditions, providing about 45K synthetic stereo images with dense disparity ground truth. In our implementation, we replace the Driving subset, in the SceneFlow dataset, with a selection of video sequences from the SYNTHIA dataset. We have chosen 10 videos sequences, covering daytime, nighttime and various seasons, providing $18,308$ training data with disparity ground truth. 

\begin{table}[t]
\centering
\resizebox{0.42\textwidth}{!}{%
\begin{tabular}{l|cc|cc|c}
\hline
\multicolumn{1}{c|}{\multirow{2}{*}{Methods}} & \multicolumn{2}{c|}{KITTI} & \multicolumn{2}{c|}{Middlebury} & \multirow{2}{*}{ETH3D} \\
\multicolumn{1}{c|}{} & 2012 & 2015 & Half & Quarter &  \\ \hline \hline
CostFilter~\cite{hosni2012fast} & 21.7 & 18.9 & 40.5 & 17.6 & 31.1 \\
PatchMatch~\cite{bleyer2011patchmatch} & 20.1 & 17.2 & 38.6 & 16.1 & 24.1 \\
SGM~\cite{hirschmuller2007stereo} & 7.1 & 7.6 & 25.2 & 10.7 & 12.9 \\ \hline
HD$^3$~\cite{yin2019hierarchical} & 23.6 & 26.5 & 37.9 & 20.3 & 54.2 \\
PSMNet~\cite{chang2018pyramid} & 27.8 & 30.7 & 34.2 & 22.7 & 16.1 \\
GwcNet~\cite{guo2019group} & 16.8 & 13.7 & 30.1 & 13.9 & 9.1 \\
GANet~\cite{zhang2019ga} & 10.1 & 11.7 & 20.3 & 11.2 & 14.1 \\ DSMNet~\cite{zhang2020domain} & 6.2 & 6.5 & 13.8 & 8.1 & 6.2 \\
Ours-PSMNet & 3.9 & \textbf{4.3} & 11.5 & 9.8 & 7.3 \\
Ours-GwcNet & \textbf{3.7} & \textbf{4.3} & \textbf{11.1} & \textbf{8.5} & \textbf{5.5} \\ \hline
\end{tabular}%
}
\caption{Evaluation of cross-domain performance using KITTI, Middlebury and ETH3D datasets. All stereo matching networks are trained using SceneFlow training sets. The results for~\cite{bleyer2011patchmatch, hirschmuller2007stereo, hosni2012fast, yin2019hierarchical, zhang2019ga} are obtained from~\cite{zhang2020domain}. \vspace{-4.5mm}}
\label{table:cross-domain}
\end{table}

We evaluate the performance of stereo disparity estimation using the commonly employed D1 error rate~(\%), with different pixel threshold. The D1 metric computes the percentage of stereo disparity outliers~(endpoint-error larger than the threshold) in the left frame. Following the advice of data originators, threshold of 3 pixels is selected for KITTI and DrivingStereo, 2 pixels for Middlebury and 1 pixel for ETH3D. \\ 

\noindent \textbf{Implementation Details: } \label{Sec:Implementation}
We have selected two popular and top-performing stereo matching networks namely PSMNet~\cite{chang2018pyramid} and GwcNet~\cite{guo2019group} as the baseline networks for our experiments. We have selected these two networks mainly due to the fact that PSMNet is well-studied, and commonly employed as a baseline in many prior works~\cite{wang2019pseudo, yao2020content, you2019pseudo, zhang2019adaptive}; and GwcNet is one of the recently proposed state-of-the-art stereo matching networks.
The proposed augmentation methods are implemented in conjunction with network architecture introduced in PSMNet and GwcNet. The networks are implemented using PyTorch framework and are trained end-to-end with Adam $(\beta_1=0.9, \beta_2=0.999)$ optimizer and smooth L1 loss function. Similar to the original implementations of PSMNet and GwcNet, our data processing includes color normalization and random cropping the input images to size $H=256$ and $W=512$. The maximum disparity is set to 192. All models are trained from scratch for 20 epochs with constant learning rate of 0.001. The batch size is set to 12 for training on 2 NVIDIA RTX 6000 Quadro GPUs. The models are trained using \textbf{synthetic data only} and directly tested using data from different realistic datasets. \\

\begin{table}[t]
\centering
\resizebox{0.38\textwidth}{!}{%
\begin{tabular}{llllllc}
\hline
\multicolumn{6}{l|}{\multirow{2}{*}{Methods}} & \multirow{2}{*}{D1 (\%)} \\
\multicolumn{6}{l|}{} &  \\ \hline
\multicolumn{7}{c}{Domain Adaptation} \\ \hline
\multicolumn{2}{l}{MADNet~\cite{tonioni2019real}} &  &  &  & \multicolumn{1}{l|}{} & 8.23 \\
\multicolumn{2}{l}{StereoGAN~\cite{liu2020stereogan}} &  &  &  & \multicolumn{1}{l|}{} & 5.74 \\
\multicolumn{2}{l}{MAML-Stereo~\cite{tonioni2019learning}} &  &  &  & \multicolumn{1}{l|}{} & 4.49 \\
\multicolumn{2}{l}{Unsupervised Adaptation~\cite{tonioni2017unsupervised}} &  &  &  & \multicolumn{1}{l|}{} & 4.02 \\ \hline
\multicolumn{7}{c}{Domain Generalization} \\ \hline
\multicolumn{2}{l}{MS-PSMNet~\cite{cai2020matchingspace}} &  &  &  & \multicolumn{1}{l|}{} & 7.76 \\
\multicolumn{2}{l}{MS-GCNet~\cite{cai2020matchingspace}} &  &  &  & \multicolumn{1}{l|}{} & 6.21 \\
\multicolumn{2}{l}{DSMNet~\cite{zhang2020domain}} &  &  &  & \multicolumn{1}{l|}{} & 4.10 \\ 
\multicolumn{2}{l}{Ours-PSMNet} &  &  &  & \multicolumn{1}{l|}{} & \textbf{4.04} \\
\multicolumn{2}{l}{Ours-GwcNet} &  &  &  & \multicolumn{1}{l|}{} & \textbf{3.83} \\ \hline
\end{tabular}%
}
\caption{Comparing performance in stereo disparity estimation with state-of-the-art domain adaptation and generalization methods using KITTI 2015 train set. \vspace{-4mm}}
\label{table:compare}
\end{table}

\begin{figure*}[ht]
    \centering
    \subfloat[Left Image]{\includegraphics[width=0.18\textwidth]{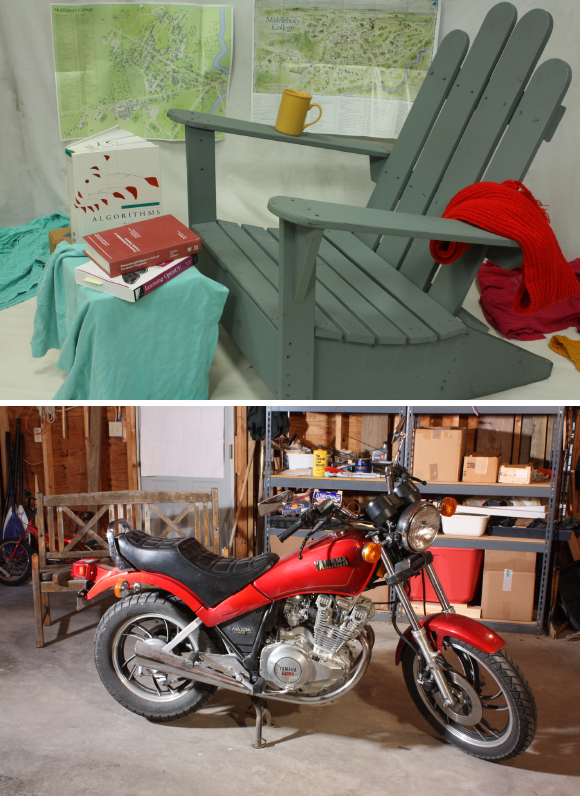}} 
    \hspace{0.3mm}
    \subfloat[PSMNet]{\includegraphics[width=0.18\textwidth]{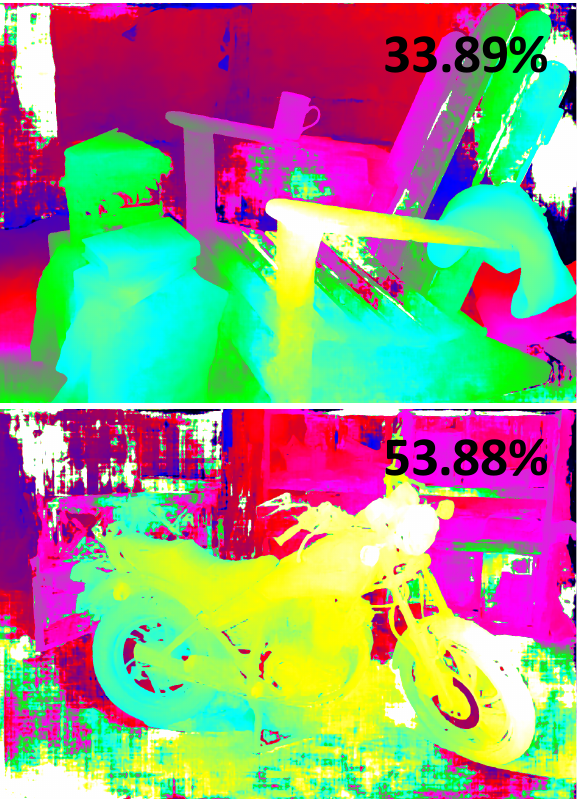}}
    \hspace{0.005mm}
    \subfloat[Ours-PSMNet]{\includegraphics[width=0.18\textwidth]{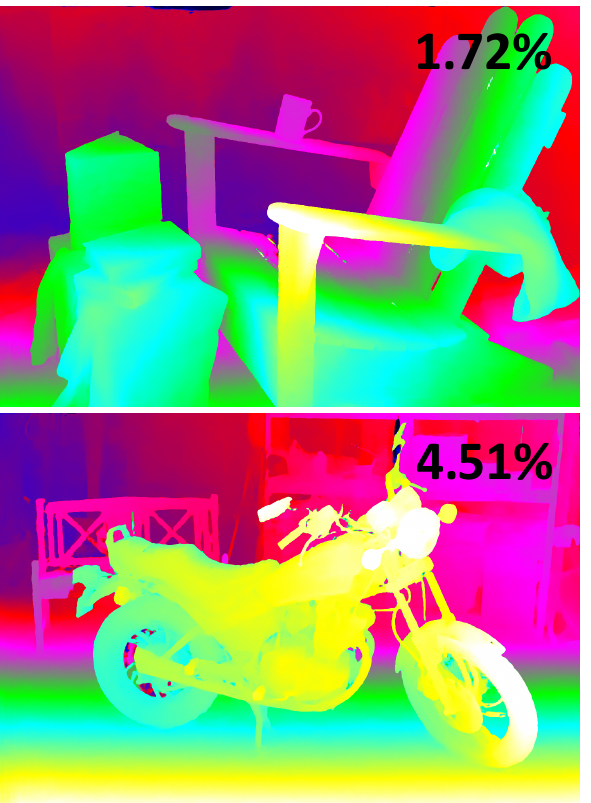}} 
    \hspace{0.005mm}
    \subfloat[GwcNet]{\includegraphics[width=0.18\textwidth]{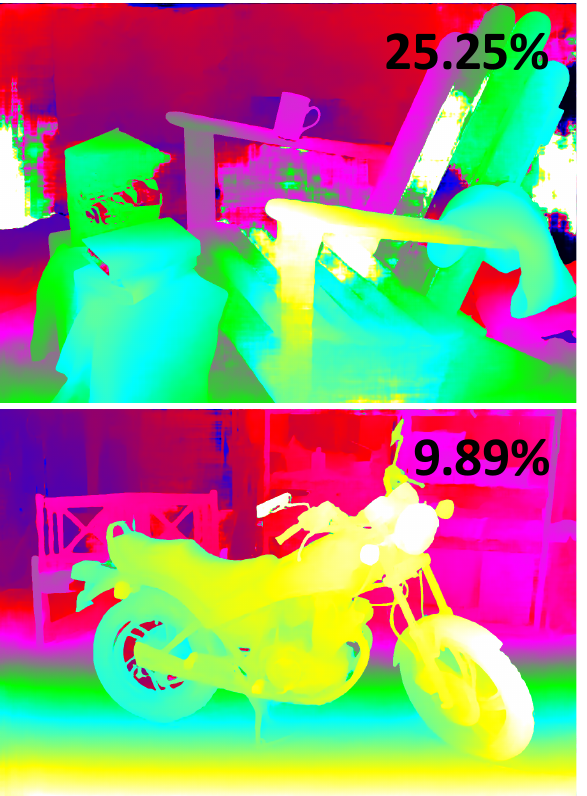}} 
    \hspace{0.005mm}
    \subfloat[Ours-Gwcnet]{\includegraphics[width=0.18\textwidth]{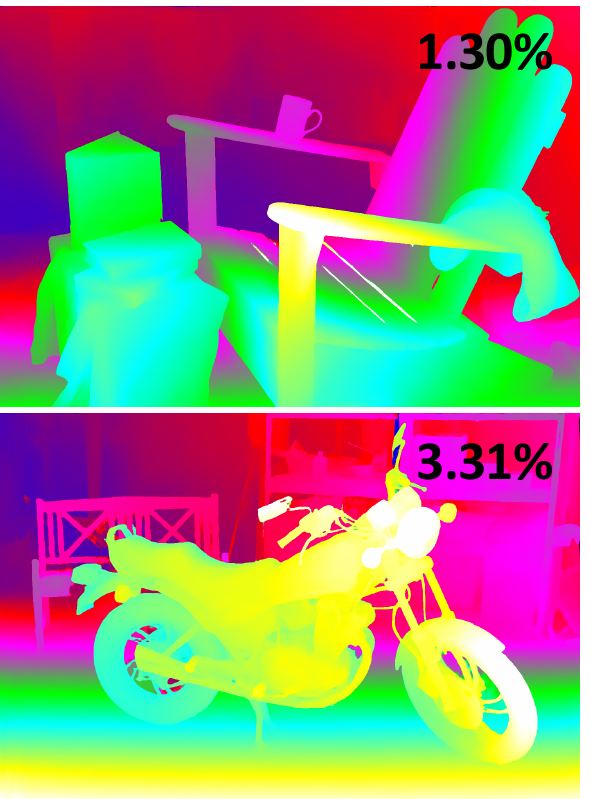}} 
    \caption{Qualitative results on Middlebury~\cite{scharstein2014high} train set for stereo matching networks~(PSMNet~\cite{chang2018pyramid} and GwcNet~\cite{guo2019group}) trained using synthetic data only, with and without the proposed data augmentation methods. The corresponding D1 error rate~(2px) is included on the top right corner of each disparity map. Additional qualitative results are included in the supplementary material.}
    \label{fig:middlebury_results}
\end{figure*}

\vspace{-2mm}
\noindent \textbf{Ablation Study: }
In this section, we evaluate the efficacy of each component~(of the proposed data augmentation methods), using multiple real datasets. As shown in Table~\ref{table:ablation}, we have achieved significant improvement in stereo disparity estimation, by including the proposed asymmetric chromatic augmentation into the learning pipeline. Moreover, the performance is further improved when the proposed asymmetric random patching augmentation is also included in the training. Remarkably, our proposed data augmentation techniques have improved the accuracy of stereo disparity estimation in KITTI~2015 dataset by about $23.9\%$ for PSMNet and about $7.5\%$ for GwcNet, despite trained using synthetic data only. Similar improvements are also observed when tested using different datasets such as Middlebury~(PSMNet:$23.5\%$, GwcNet:$16.4\%$) and ETH3D~(PSMNet:$4.1\%$, GwcNet:$4.4\%$). Qualitative comparisons on Middlebury are included in Fig.~\ref{fig:middlebury_results}. \\

\noindent \textbf{Component Analysis and Comparisons: }
We further validate the effectiveness of our proposed data augmentation techniques in generalizing stereo matching networks by comparing with domain normalization~(DN) proposed by Zhang~\etal~\cite{zhang2020domain}. While DN can effectively normalize the distributions of learned features and reduces the impact of domain shifting, in the learning process, the issues associated with the attributes of synthetic data still persist.  In Table~\ref{table:DN}, our methods outperform DN when tested with two different networks, validating that eliminating shortcut learning in stereo matching networks is key to achieve domain-invariant generalization. \\

\vspace{-8mm}
\begin{table}[ht]
\centering
\resizebox{0.4\textwidth}{!}{%
\begin{tabular}{l||cccc}
\hline
\multirow{2}{*}{Weather} & \multicolumn{4}{c}{D1 (\%)} \\ \cline{2-5} 
 & Sunny & Cloudy & Rainy & Foggy \\ \hline
GwcNet~\cite{guo2019group} & \textbf{3.39} & \textbf{2.76} & 12.33 & 7.12 \\
Ours-GwcNet & 3.75 & 3.34 & \textbf{8.53} & \textbf{6.92} \\ \hline
\end{tabular}%
}
\caption{Evaluation of network robustness using DrivingStereo datasets collected in different weather conditions. GwcNet was pre-trained using SceneFlow data and fine-tuned on KITTI2015 dataset. Meanwhile, ours-GwcNet was trained on synthetic data only, using the proposed data augmentation methods.}
\label{table:weather}
\end{table} 

\noindent \textbf{Cross Domain Analysis: }
In this section, we evaluate the cross domain generalization of our methods using three datasets. Following the footstep of Zhang~\etal~\cite{zhang2020domain}, we also compare our results with the traditional stereo matching algorithm including SGM~\cite{hirschmuller2007stereo}, PatchMatch~\cite{bleyer2011patchmatch} and CostFilter~\cite{hosni2012fast}; as well as with the state-of-the-art stereo matching networks~\cite{chang2018pyramid, guo2019group, yin2019hierarchical, zhang2019ga}. All networks are trained on SceneFlow synthetic data only. As shown in Table~\ref{table:cross-domain}, our methods significantly improve the generalization of PSMNet and GwcNet, outperforming the state-of-the-art networks and the traditional stereo matching algorithms in all indoor and outdoor datasets. 

In addition, we also compare our methods with other state-of-the-art domain generalizing and domain adaptation in the context of stereo matching networks. As illustrated in Table~\ref{table:compare} we have achieved the best performance in stereo disparity estimation without using any data from the target domain~(KITTI 2015 dataset). \\

\begin{figure*}[ht]
    \centering
    \includegraphics[width=0.95\textwidth]{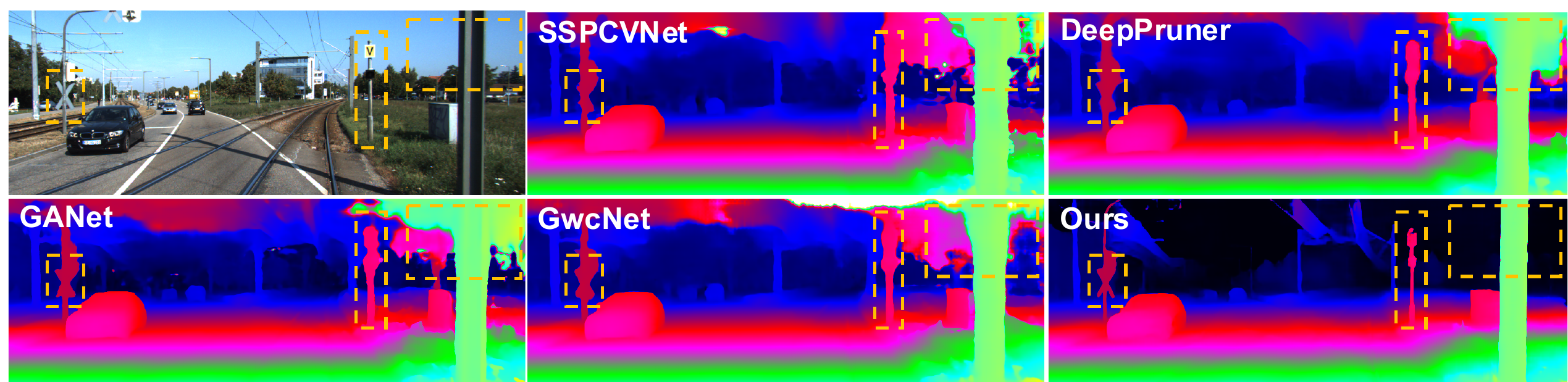}
    \caption{Qualitative results on KITTI~2015 benchmark. By eliminating the identified shortcuts using data augmentations, the resulting stereo matching network can accurately estimate disparities for thin objects and top-half of the image, despite being trained using synthetic data only. Yellow-dashed boxes indicate the improved areas where our method can generate reliable disparity estimates with sharp object boundaries as compared to the state-of-the-arts networks that are fine-tuned on KITTI domain. Additional qualitative results are included in the supplementary material.}
    \label{fig:kitti_benchmark}
\end{figure*}

\begin{table}[t]
\centering
\resizebox{0.38\textwidth}{!}{%
\begin{tabular}{l|c|cc}
\hline
\multirow{2}{*}{Methods} & \multicolumn{1}{l|}{\multirow{2}{*}{Training Set}} & \multicolumn{2}{c}{D1 (\%)} \\
 & \multicolumn{1}{l|}{} & All & Noc \\ \hline \hline
DispNetC~\cite{mayer2016large} & Kitti-gt & 4.34 & 4.05 \\
Content-CNN~\cite{luo2016efficient} & Kitti-gt & 4.54 & 4.00 \\
MADNet-ft~\cite{tonioni2019real} & Kitti-gt & 4.66 & 4.27 \\ \hline
DispSegNet~\cite{zhang2019dispsegnet} & Kitti & 6.33 & 5.85 \\
SegStereo~\cite{yang2018segstereo} & Kitti & 8.79 & 7.70 \\
OASM-Net~\cite{li2018occlusion} & Kitti & 8.98 & 7.39 \\
Unsupervised~\cite{zhou2017unsupervised} & Kitti & 9.91 & 8.61 \\ \hline
Ours-PSMNet & Synthetic & 4.27 & 4.02 \\
Ours-GwcNet & Synthetic & \textbf{4.11} & \textbf{3.83} \\ \hline
\end{tabular}%
}
\caption{Performance comparison with supervised and unsupervised learning-based stereo matching networks on KITTI~2015 online benchmark. \vspace{-4mm}} 
\label{table:kitti15-benchmark}
\end{table}

\noindent \textbf{Network Robustness Analysis: }
In this section, we empirically show that the proposed data augmentation methods can effectively constrain the stereo matching networks~(to learn \textbf{robust} features), we evaluate the performance of GwcNet trained under two settings: (1) pre-trained on SceneFlow and fine-tuned on KITTI 2015 dataset, without data augmentations~(baseline), and (2) trained on synthetic data only~(SceneFlow + SYNTHIA) and with the proposed data augmentations included~(Ours). The trained networks are tested using data collected in different weather conditions provided by DrivingStereo and the results are summarized in Table~\ref{table:weather}. As the DrivingStereo and KITTI 2015 datasets consist of similar outdoor driving scenes, the network trained on KITTI can generalize well to DrivingStereo. However, as the KITTI 2015 dataset only consists of images collected in ideal weather conditions~(sunny and cloudy), the baseline network fails when tested with data collected in adverse weather conditions~(rain and fog). 

\begin{figure}[t]
    \centering
    \subfloat[Left Image]{\includegraphics[width=0.115\textwidth]{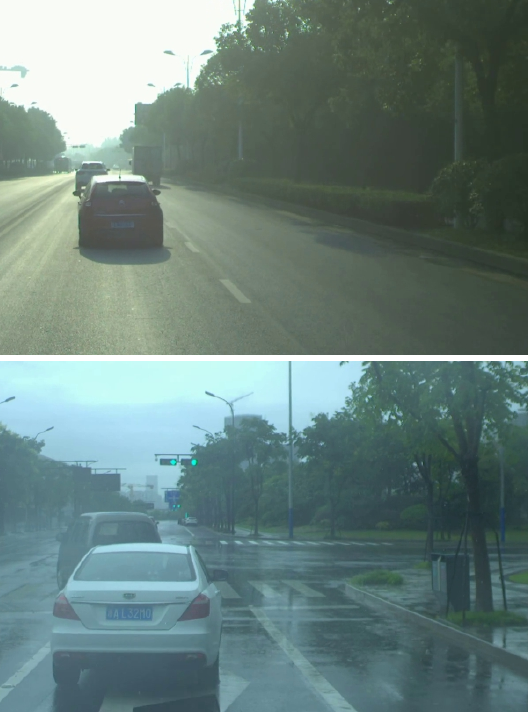}} 
    \hspace{0.02mm}
    \subfloat[Right Image]{\includegraphics[width=0.115\textwidth]{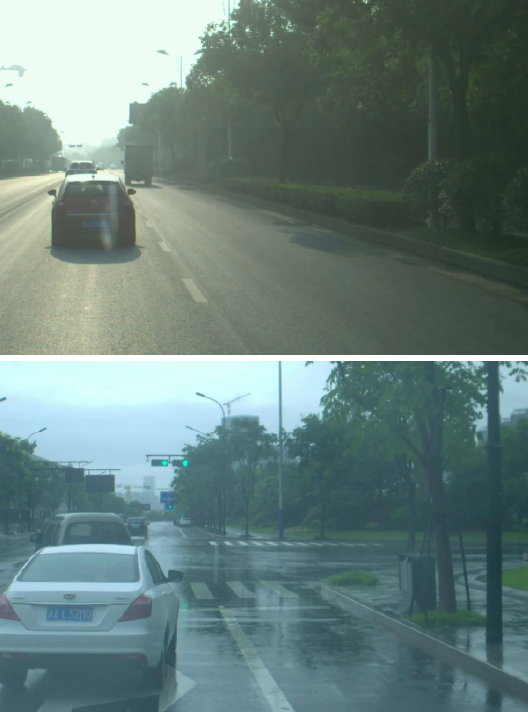}}
    \hspace{0.01mm}
    \subfloat[Baseline]{\includegraphics[width=0.115\textwidth]{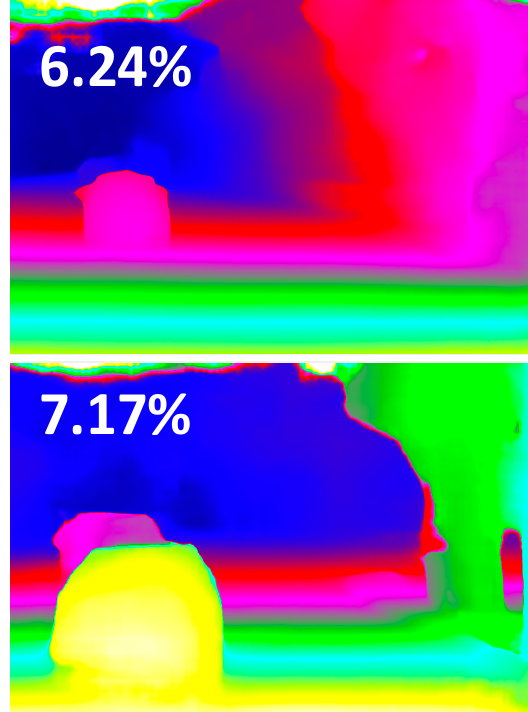}}
    \hspace{0.01mm}
    \subfloat[Ours]{\includegraphics[width=0.115\textwidth]{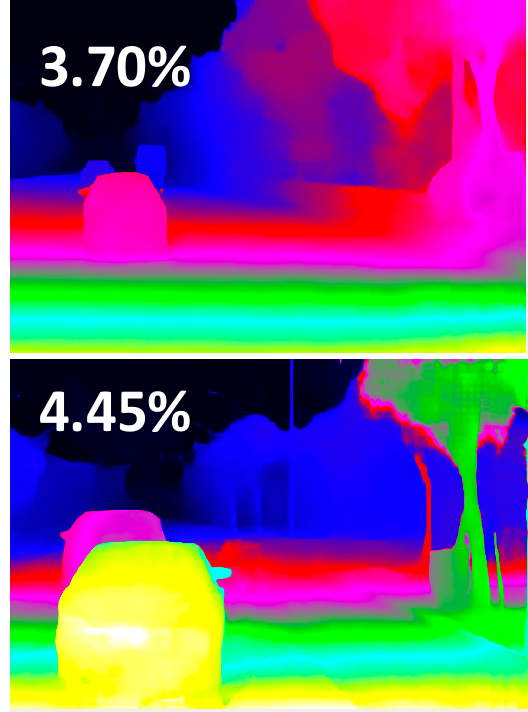}}
    \caption{Qualitative analysis of GwcNet trained with different settings, using challenging examples collected in adverse weather conditions: foggy~(top) and rainy~(bottom). The corresponding D1 error rate~(3 pixels) is also included on the top left corner of each estimated disparity map.} 
    \label{fig:weather}
\end{figure}

In contrast, our method performs significantly better than the baseline in adverse weather conditions while achieving comparable performance in ideal weathers, even though trained using \textbf{synthetic data only}. Qualitative comparisons are included in Fig.~\ref{fig:weather}. 
The results validate our hypothesis -- eliminating shortcut learning from stereo matching networks using data augmentations can result in the learning of \textbf{robust features} and achieving domain-invariant generalization. \\

\noindent \textbf{KITTI Benchmark Evaluation: }
We also submitted our results to the KITTI 2015 online benchmark to compare against prior works. In Table~\ref{table:kitti15-benchmark}, our models, despite trained on synthetic data only, significantly outperform most of the unsupervised models that are trained on a large set of KITTI stereo samples. Impressively, our methods also achieve a lower error rate than some of the supervised stereo matching networks that are fine-tuned using the labelled KITTI dataset.

Moreover, as highlighted in Fig.~\ref{fig:kitti_benchmark}, our methods can produce sharp boundaries and accurate shape representation for most objects compared to the state-of-the-art stereo matching networks~(GANet~\cite{zhang2019ga}, GwcNet~\cite{guo2019group}, DeepPruner~\cite{Duggal2019ICCV}, SSPCVNet~\cite{Wu_2019_ICCV}) that are fine-tuned on KITTI training data. In addition, our methods can also estimate accurate disparity measurements for the upper half area, while the mentioned state-of-the-arts methods usually fail in this aspect~(as shown in Fig.~\ref{fig:kitti_benchmark}). 
\vspace{-1mm}
\section{Conclusion}~
This paper shows that eliminating the identified shortcuts~(posed by the non-photorealistic synthetic training data) is an important key to achieve domain robustness in stereo matching networks. We showed that the commonly used synthetic training data have the following issues: (1)~identical color features between matching pixels and (2)~lack of realistic textures on synthetic objects. These attributes result in the stereo matching network learning shortcuts that create superior performance for synthetic data, which does not translate to success with realistic data. To this end, we advocate eliminating the shortcut learning opportunities, by augmenting the synthetic data using asymmetric chromatic augmentation and asymmetric random patching and showed that including those can effectively prevent networks from relying on the identical color features of synthetic data. Moreover, we demonstrated that using asymmetric random patching can constrain the networks to incorporate global features: further improving the domain generalization. Our experimental results illustrated that the proposed methods effectively promoted domain invariant generalization and network robustness, thus achieving a substantial improvement in stereo disparity estimation on multiple challenging realistic datasets while networks were only trained using synthetic data. This work provides a significant step towards generating reliable and robust stereo depth estimation systems, while utilizing synthetic data only. This can be beneficial to many real-world applications that require accurate depth measuring system.

{\small
\bibliographystyle{ieee_fullname}
\bibliography{main}
}

\end{document}